\documentclass[11pt,a4paper]{article}
\usepackage[hyperref]{acl2018}
\usepackage{times}
\usepackage{latexsym}
\usepackage{url}
\usepackage{soul}
\usepackage{multirow}
\usepackage{tabularx}
\usepackage{arydshln}
\usepackage{tikz}
\usepackage{pgfplots}
\usepackage{array}
\usepackage[export]{adjustbox}

\aclfinalcopy %
\def\aclpaperid{1194} %

\usepackage{amssymb}
\usepackage{amsmath}
\usepackage{graphicx}
\usepackage{balance}

\usepackage{titlesec}
\usepackage{caption}
\DeclareCaptionFormat{capfont}{\fontsize{10}{6}\selectfont#1#2#3}
\usepackage{subcaption}
\makeatletter
\g@addto@macro\small{%
  \setlength\abovedisplayskip{-5pt}
  \setlength\abovedisplayshortskip{-5pt}
  \setlength\belowdisplayshortskip{-7pt}
  \setlength\belowdisplayskip{-7pt}
}
\makeatother

\usepackage{algorithmicx} 
\usepackage[noend]{algpseudocode}
\usepackage{algorithm}
\usepackage[T1]{fontenc}

\algnewcommand\algorithmicdefinitions{\textbf{Definitions:}}
\algnewcommand\Definitions{\item[\algorithmicdefinitions]}
\renewcommand{\algorithmiccomment}[1]{{\color{gray}\raisebox{1px}{\texttt{\guillemotright}} #1}}
\algnewcommand{\LineComment}[1]{\Statex \hskip\ALG@thistlm \algorithmiccomment{#1}}
\algrenewcommand\alglinenumber[1]{\footnotesize #1:}
\algrenewcommand\algorithmicindent{1.0em}%
\makeatletter
\newcommand{\StatexIndent}[1][3]{%
  \setlength\@tempdima{\algorithmicindent}%
  \Statex\hskip\dimexpr#1\@tempdima\relax}

\newcommand{\dline}{\hdashline[0.5pt/1pt]}
\newcommand{\ddline}[1]{\cdashline{#1}[0.5pt/1pt]}

\newcommand{\eat}[1]{}

\newcommand{\nlstring}[1]{{\em #1}}
\newcommand{\const}[1]{{\rm #1}}

\newcommand{\length}[1]{|{#1}|}
\newcommand{\func}[1]{\text{\MakeUppercase{#1}}}

\newcommand{\inputtoken}{x}

\newcommand{\action}{a}
\newcommand{\act}[1]{{\tt \MakeUppercase{#1}}}

\newcommand{\instruction}{\bar{\inputtoken}}
\newcommand{\state}{s}
\newcommand{\acontext}{\tilde{s}}

\newcommand{\goalstate}{g}

\newcommand{\execution}{\bar{e}}
\newcommand{\transition}{T}
\newcommand{\policy}{\pi}

\newcommand{\interactionlength}{n}

\newcommand{\outputlength}{m}

\newcommand{\states}{\mathcal{S}}
\newcommand{\actions}{\mathcal{A}}
\newcommand{\instructions}{\mathcal{X}}

\newcommand{\turnindex}{i}
\newcommand{\inputindex}{j}
\newcommand{\outputindex}{k}
\newcommand{\dataindex}{{(j)}}

\newcommand{\beaker}{\bar{b}}
\newcommand{\alchemylength}{N}
\newcommand{\beakercolor}{c}

\newcommand{\beakerindex}{i}

\newcommand{\scenestate}{S}
\newcommand{\spot}{p}
\newcommand{\scenelength}{N}
\newcommand{\hatcolor}{h}
\newcommand{\shirtcolor}{s}
\newcommand{\spotindex}{i}

\newcommand{\tangramstate}{T}
\newcommand{\tangramlength}{n}
\newcommand{\shapeembedding}{\embedding^s}

\newcommand{\embedding}{\phi}
\newcommand{\rnn}{\func{LSTM}}
\newcommand{\hiddenstate}{\mathbf{h}}

\newcommand{\inembedding}{\embedding^I}
\newcommand{\unitembedding}{\embedding^c}
\newcommand{\posembedding}{\embedding^p}
\newcommand{\outputembedding}{\embedding^O}

\newcommand{\encoderrnn}{\rnn^E}
\newcommand{\beakerrnn}{\rnn^B}
\newcommand{\scenernn}{\rnn^S}
\newcommand{\decoderrnn}{\rnn^D}

\newcommand{\stdev}[2]{${#1}$\begin{tiny}$\pm{#2}$\end{tiny}}

\title{Situated Mapping of Sequential Instructions to Actions \\ with Single-step Reward Observation}

\author{Alane Suhr {\normalfont and} Yoav Artzi\\
   Department of Computer Science and Cornell Tech\\
 Cornell University \\
  New York, NY, 10044 \\
  {\tt \{suhr, yoav\}@cs.cornell.edu}}

\date{}

\begin{document}

\maketitle
\begin{abstract}
We propose a learning approach for mapping context-dependent sequential instructions to actions. We address the problem of discourse and state dependencies with an attention-based model that considers both the history of the interaction and the state of the world. To train from start and goal states without access to demonstrations, we propose SESTRA, a learning algorithm that takes advantage of single-step reward observations and immediate expected reward maximization. We evaluate on the SCONE domains, and show absolute accuracy improvements of 9.8\%-25.3\% across the domains over approaches that use high-level logical representations.

\end{abstract}

\section{Introduction}
\label{sec:intro}

\begin{figure}[t]
\fbox{
\centering
\begin{minipage}{0.95\linewidth}
	\begin{center}
	\includegraphics[width=0.7\linewidth,clip,trim=112 425 567 167]{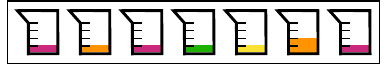} \\		
	\end{center}
	\vspace{-8pt}
	\footnotesize
	\nlstring{throw out first beaker} \\[1pt]
	$\act{pop~1}, \act{stop}$\\[3pt]
	\nlstring{pour sixth beaker into last one} \\[1pt]
	$\act{pop~6}, \act{pop~6}, \act{push~7~o}, \act{push~7~o}, \act{stop}$ \\[3pt]
	\nlstring{it turns brown} \\[1pt]
	$\act{pop~7}, \act{pop~7}, \act{pop~7}, \act{push~7~b}, \act{push~7~b}, \act{push~7~b}, \act{stop}$ \\[3pt]
	\nlstring{pour purple beaker into yellow one} \\[1pt]
	$\act{pop~3}, \act{push~5~p}, \act{stop}$\\[3pt]
	\nlstring{throw out two units of brown one} \\[1pt]
	$\act{pop~7}, \act{pop~7}, \act{stop}$
	\begin{center}
	\vspace{-6pt}
	\includegraphics[width=0.7\linewidth,clip,trim=112 371 567 211]{figs/start-goal} \\		
	\end{center}
\end{minipage}}
\vspace{-5pt}
\caption{Example from the SCONE~\cite{Long:16context} \textsc{Alchemy} domain, including a start state (top), sequence of instructions, and a goal state (bottom). Each instruction is annotated with a sequence of actions from the set of actions we define for \textsc{Alchemy}.}
\label{fig:example}	
\vspace{-10pt}
\end{figure}

An agent executing a sequence of instructions must address multiple challenges, including grounding the language to its observed environment, reasoning about discourse dependencies, and generating actions to complete high-level goals. 
For example, consider the environment and instructions in Figure~\ref{fig:example}, in which a user describes moving chemicals between beakers and mixing chemicals together. 
To execute the second instruction, the agent needs to resolve \nlstring{sixth beaker} and \nlstring{last one} to objects in the environment. 
The third instruction requires resolving \nlstring{it} to the rightmost beaker mentioned in the second instruction, and reasoning about the set of actions required to mix the colors in the beaker to brown. 
In this paper, we describe a model and learning approach to map sequences of instructions to actions. 
Our model considers previous utterances and the world state to select actions, learns to combine simple actions to achieve complex goals, and can be trained using goal states without access to demonstrations.

The majority of work on executing sequences of instructions focuses on mapping instructions to high-level formal representations, which are then evaluated to generate actions~\cite[e.g.,][]{Chen:11,Long:16context}.
For example, the third instruction in Figure~\ref{fig:example} will be mapped to  $\const{mix}(\const{prev\_arg1})$, indicating that the mix action should be applied to first argument of the previous action~\cite{Long:16context,Guu:17rl-mml}. 
In contrast, we focus on directly generating the sequence of actions. 
This requires resolving references without explicitly modeling them, and  learning the sequences of actions required to complete high-level actions; for example, that mixing requires removing everything in the beaker and replacing with the same number of brown items.

A key challenge in executing sequences of instructions is considering contextual cues from both the history of the interaction and the state of the world. 
Instructions often refer to previously mentioned objects (e.g., \nlstring{it} in Figure~\ref{fig:example}) or actions (e.g., \nlstring{do it again}). 
The world state provides the set of objects the instruction may refer to, and implicitly determines the available actions. For example, liquid can not be removed from an empty beaker. 
Both types of contexts continuously change during an interaction. As new instructions are given, the instruction history expands, and as the agent acts the world state changes. 
We propose an attention-based model that takes as input the current instruction, previous instructions, the initial world state, and the current state. 
At each step, the model computes attention encodings of the different inputs, and predicts the next action to execute. 

We train the model given instructions paired with start and goal states without access to the correct sequence of actions. 
During training, the agent learns from rewards received through exploring the environment with the learned policy  by mapping instructions to sequences of actions. 
In practice, the agent learns to execute instructions gradually, slowly correctly predicting prefixes of the correct sequences of increasing length  as learning progress. 
A key  challenge  is learning to correctly select actions that are only required later in execution sequences. 
Early during learning, these actions receive negative updates, and the agent learns to assign them low probabilities. 
This results in an exploration problem in later stages, where actions that are only required later are not sampled during exploration. 
For example, in the \textsc{Alchemy} domain shown in Figure~\ref{fig:example}, the agent behavior early during execution of instructions can be accomplished  by only using $\act{pop}$ actions. 
As a result, the agent quickly learns a strong bias against $\act{push}$ actions, which in practice prevents the policy from exploring them again.   
We address this with a learning algorithm that  observes the reward for all possible actions for each visited state, and maximizes the immediate expected reward.

We evaluate our approach on SCONE~\cite{Long:16context}, which includes three domains, and is used to study recovering predicate logic meaning representations for sequential instructions. 
We study the problem of generating a sequence of low-level actions, and re-define the set of actions for each domain. 
For example, we treat the beakers in the \textsc{Alchemy} domain as stacks and use only $\act{pop}$ and $\act{push}$ actions. 
Our approach robustly learns to execute sequential instructions with up to $89.1\%$ task-completion accuracy for single instruction, and  $62.7\%$ for complete sequences. 
Our code is available at \mbox{\url{https://github.com/clic-lab/scone}}.

\section{Technical Overview}
\label{sec:overview}

\paragraph{Task and Notation}

Let $\states$ be the set of all possible world states, $\instructions$ be the set of all natural language instructions, and $\actions$ be the set of all actions.
An instruction $\instruction \in \instructions$ of length $\length{\instruction}$ is a sequence of tokens $\langle \inputtoken_1, ... \inputtoken_{\length{\instruction}} \rangle$. 
Executing an action modifies the world state following a transition function $\transition : \states \times \actions \rightarrow \states$. For example, the \textsc{Alchemy} domain includes seven beakers that contain colored liquids. The world state defines the content of each beaker. We treat each beaker as a stack. The actions are $\act{pop~N}$ and $\act{push~N~C}$, where $ 1 \leq \act{N} \leq 7$ is the beaker number and $\act{C}$ is one of six colors. There are a total of $50$ actions, including the $\act{stop}$ action. Section~\ref{sec:data} describes the domains in detail.

Given a start state $\state_1$ and a sequence of instructions $\langle \instruction_1, \dots, \instruction_\interactionlength \rangle$, our goal is to generate the sequence of actions specified by the instructions starting from $\state_1$. 
We treat the execution of a sequence of instructions as executing each instruction in turn. 
The execution $\execution$ of an instruction $\instruction_\turnindex$ starting at a state $\state_1$ and given the history of the instruction sequence $\langle \instruction_1, \dots, \instruction_{\turnindex-1}\rangle$ is a sequence of state-action pairs $\execution = \langle (\state_1, \action_1), ..., (\state_\outputlength, \action_\outputlength) \rangle$, where  $\action_\outputindex \in \actions$, $\state_{\outputindex+1} = \transition (\state_\outputindex, \action_\outputindex)$. 
The final action $\action_\outputlength$ is the special action $\act{STOP}$, which indicates the execution has terminated. 
The final state is then $\state_\outputlength$, as $\transition(\state_\outputindex, \mathtt{STOP}) = \state_\outputindex$.
Executing a sequence of instructions in order generates a sequence $\langle \execution_1, ..., \execution_\interactionlength \rangle$, where $\execution_\turnindex$ is the execution of instruction $\instruction_\turnindex$.
When referring to states and actions in an indexed execution $\execution_\turnindex$,  the $\outputindex$-th state and action are $\state_{\turnindex, \outputindex}$ and $\action_{\turnindex, \outputindex}$. 
We execute instructions one after the other: $\execution_1$ starts at the interaction initial state $\state_1$ and $\state_{\turnindex+ 1,1} = \state_{\turnindex,\length{\execution_\turnindex}}$, where $\state_{\turnindex + 1,1}$ is the start state of $\execution_{\turnindex+1}$ and $\state_{\turnindex, \length{\execution_\turnindex}}$ is the final state of $\execution_\turnindex$.

\paragraph{Model}
We model the agent with a neural network policy (Section~\ref{sec:model}). 
At step $\outputindex$ of executing the $\turnindex$-th instruction, the model input is the current instruction $\instruction_\turnindex$, the previous instructions $\langle \instruction_1,\dots, \instruction_{\turnindex- 1}\rangle$, the world state $\state_1$ at the beginning of executing  $\instruction_\turnindex$, and the current state $\state_{\outputindex}$. 
The model predicts the next action $\action_{\outputindex}$ to execute. 
If $\action_{\outputindex} = \act{stop}$, we switch to the next instruction, or if at the end of the instruction sequence, terminate. 
Otherwise, we update the state to $\state_{\outputindex+1} = \transition(\state_{ \outputindex}, \action_{\outputindex} )$. 
The model uses attention to process the different inputs and a recurrent neural network (RNN) decoder to generate actions~\cite{Bahdanau:14neuralmt}.

\paragraph{Learning}

We assume access to a set of $N$ instruction sequences, where each instruction in each sequence is paired with its start and goal states. 
During training, we create an example for each instruction. 
Formally, the training set is $\{(\instruction^\dataindex_\turnindex, \state_{\turnindex, 1}^\dataindex, \langle \instruction^\dataindex_1, \dots, \instruction^\dataindex_{\turnindex-1} \rangle, \goalstate^\dataindex_\turnindex) \}_{j=1,\turnindex=1}^{N,\interactionlength^\dataindex}$, where $\instruction^\dataindex_\turnindex$ is an instruction, $\state_{\turnindex, 1}^\dataindex$ is a start state, $\langle \instruction^\dataindex_1, \dots, \instruction^\dataindex_{\turnindex-1} \rangle$ is the instruction history, $\goalstate^\dataindex_\turnindex$ is the goal state, and $\interactionlength^\dataindex$ is the length of the  $j$-th instruction sequence. 
This training data contains no evidence about the actions and intermediate states required to execute each instruction.\footnote{This training set is a subset of the data used in previous work~\cite[Section 6,][]{Guu:15:traverse}, in which training uses all instruction  sequences of length $1$ and $2$.} 
We use a learning method that maximizes the expected immediate reward for a given state (Section~\ref{sec:learning}). 
The reward accounts for task-completion and distance to the goal via potential-based reward shaping.

\paragraph{Evaluation}

We evaluate exact task completion for sequences of instructions on a test  set $\{(\state^\dataindex_1, \langle \instruction_1^\dataindex, \dots,\instruction_{\interactionlength_j}^\dataindex\rangle, \goalstate^\dataindex)\}_{j=1}^N$, where $\goalstate^\dataindex$ is the oracle goal state of executing instructions \mbox{$\instruction_1^\dataindex$, \dots,$\instruction_{\interactionlength_j}^\dataindex$} in order starting from $\state^\dataindex_1$. 
We also evaluate single-instruction task completion using per-instruction annotated start and goal states.

\section{Related Work}
\label{sec:related}

Executing instructions has been studied using the SAIL corpus~\cite{MacMahon:06} with focus on navigation using high-level logical representations~\cite{Chen:11,Chen:12,Artzi:13,Artzi:14} and low-level actions~\cite{Mei:16neuralnavi}. 
While SAIL includes sequences of instructions, the data demonstrates limited discourse phenomena, and instructions are often processed in isolation. 
Approaches that consider as input the entire sequence focused on segmentation~\cite{Andreas:15navi}. 
Recently, other navigation tasks were proposed with focus on single instructions~\cite{Anderson:17,Janner:17spatialrep}. 
We focus on sequences of environment manipulation instructions and modeling contextual cues from both the changing environment and instruction history. 
Manipulation using single-sentence instructions has been studied using the Blocks domain~\cite{Bisk:16nl-robots,Bisk:17,Misra:17instructions,Tan:17blocks}. 
Our work is related to the work of \citet{Branavan:09} and \citet{Vogel:10}. 
While both study executing sequences of instructions, similar to SAIL, the data includes limited discourse dependencies. 
In addition, both learn with rewards computed from surface-form similarity between text in the environment and the instruction.  
We do not rely on such similarities, but instead use a state distance metric. 
Language understanding in interactive scenarios that include multiple turns has been studied  with focus on dialogue for querying database systems using the ATIS corpus~\cite{Hemphill:90atis,Dahl:94}.  
\citet{Tur:10atis} surveys work on ATIS. 
\citet{Miller:96}, \citet{Zettlemoyer:09}, and \citet{Suhr:18atis} modeled context dependence in ATIS for generating formal representations.
In contrast, we focus on environments that change during execution and directly generating environment actions, a scenario that is more related to robotic agents than database query. 
The SCONE corpus~\cite{Long:16context} was designed to reflect a broad set of discourse context-dependence phenomena. 
It was studied extensively using logical meaning representations~\cite{Long:16context,Guu:17rl-mml,Fried:17}.  
In contrast, we are interested in directly generating actions that modify the environment. This requires generating lower-level actions and learning procedures that are otherwise hardcoded in the logic (e.g., mixing action in Figure~\ref{fig:example}). 
Except for \citet{Fried:17}, previous work on SCONE assumes access only to the initial and final states during training. 
This form of supervision does not require operating the agent manually to acquire the correct sequence of actions, a difficult task in robotic agents with complex control. 
Goal state supervision has been studied for instructional language~\cite[e.g.,][]{Branavan:09,Artzi:13,Bisk:16nl-robots}, and more extensively in question answering when learning with answer annotations only~\cite[e.g.,][]{Clarke:10,Liang:11,Kwiatkowski:13,Berant:13,Berant:14paraphrasing,Berant:15imitation,Liang:17freebase}.

\section{Model}
\label{sec:model}

\begin{figure*}
	\centering
	\includegraphics[width=\textwidth,clip,trim=70 311 273 163]{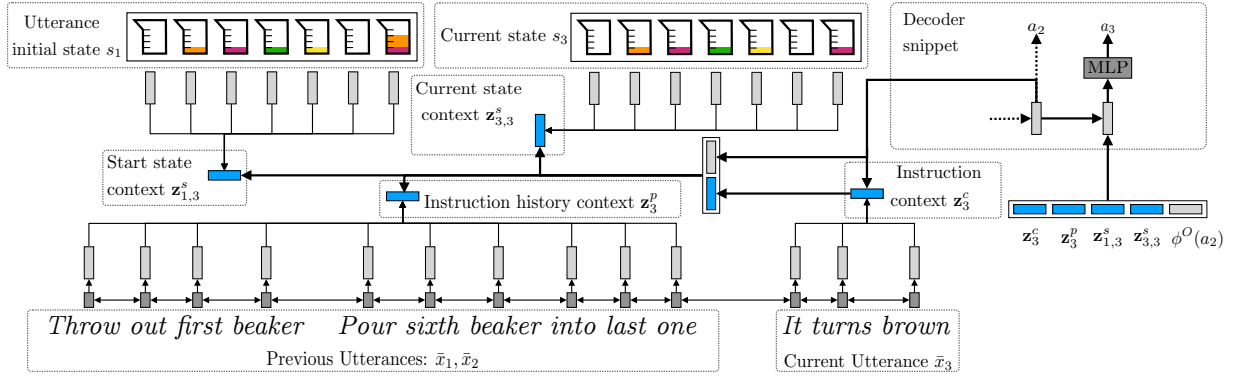}
	\vspace{-15pt}
	\caption{Illustration of the model architecture while generating the third action $\action_3$ in the third utterance $\instruction_3$ from Figure~\ref{fig:example}. Context vectors computed using attention are highlighted in blue. The model takes as input vector encodings from the current and previous instructions $\instruction_1$, $\instruction_2$, and $\instruction_3$, the initial state $\state_1$, the current state $\state_3$, and the previous action $\action_2$. Instruction encodings are computed with a bidirectional RNN. We attend over the previous and current instructions and the initial and current states. We use an MLP to select the next action.}
	\label{fig:arch}
	\vspace{-15pt}
\end{figure*}

\newcommand{\instructionencodings}{X}
\newcommand{\stateencoding}{S}
\newcommand{\contextvector}{\mathbf{z}}

We map sequences of instructions $\langle \instruction_1, \dots, \instruction_\interactionlength \rangle$ to actions by executing the instructions in order. 
The model generates an execution $\execution = \langle (\state_1, \action_1), \dots, (\state_{\outputlength_\turnindex}, \action_{\outputlength_\turnindex})\rangle$ for each instruction $\instruction_\turnindex$.
The agent context, the information available to the agent at step $\outputindex$, is $\acontext_\outputindex = (\instruction_\turnindex, \langle \instruction_1, \dots, \instruction_{\turnindex-1}\rangle, \state_\outputindex, \execution[:\outputindex])$, where $\execution[:\outputindex]$ is the execution up until but not including step $k$. 
In contrast to the world state, the agent context also includes instructions and the execution so far. 
The agent policy $\policy_\theta(\acontext_\outputindex, \action)$ is modeled as a probabilistic neural network parametrized by $\theta$, where $\acontext_\outputindex$ is the agent context at step $\outputindex$ and $\action$ is an action. 
To generate executions, we generate one action at a time, execute the action, and observe the new world state. 
In step $\outputindex$ of executing the  $\turnindex$-th instruction, the network inputs are the current utterance $\instruction_\turnindex$, the previous instructions $\langle \instruction_1, \dots, \instruction_{\turnindex-1}\rangle$, the initial state $\state_1$ at beginning of executing $\instruction_\turnindex$, and the current state $\state_\outputindex$. 
When executing a sequence of instructions, the initial state $\state_1$ is either the state at the beginning of executing the sequence or the final state of the execution of the previous instruction. 
Figure~\ref{fig:arch} illustrates our architecture.

We generate continuous vector representations for all inputs. 
Each input is represented as a set of vectors that are then processed with an attention function to generate a single vector representation~\cite{Luong:15nmtattention}. 
We assume access to a domain-specific encoding function $\func{enc}(\state)$ that, given a state $\state$, generates a set of vectors $\stateencoding$ representing the objects in the state. 
For example, in the \textsc{Alchemy} domain, a vector is generated for each beaker using an RNN. 
Section~\ref{sec:data} describes the different domains and their encoding functions.

We use a single bidirectional RNN with a long short-term memory~\cite[LSTM;][]{Hochreiter:97lstm} recurrence to encode the instructions. 
All instructions $\instruction_1$,\dots,$\instruction_\turnindex$ are encoded with a single RNN by concatenating them to $\instruction'$. 
We use two delimiter tokens: one separates previous instructions, and the other separates the previous instructions from the current one.
The forward LSTM RNN hidden states are computed as:\footnote{To simplify the notation, we omit the memory cell (often denoted as $\mathbf{c}_\inputindex$) from all LSTM descriptions. We use only the hidden state $\hiddenstate_\inputindex$ to compute the intended representations (e.g., for the input text tokens). All LSTMs in this paper use zero vectors as initial hidden state $\hiddenstate_0$ and initial cell memory $\mathbf{c}_0$.}

\begin{small}
\begin{equation}
\nonumber \overrightarrow{\hiddenstate_{j+1}} = \overrightarrow{\encoderrnn}\left(\inembedding(\inputtoken'_{j+1});  \overrightarrow{\hiddenstate_j}\right)\;\;,
\end{equation}
\end{small}

\noindent
where $\inembedding$ is a learned word embedding function and $\overrightarrow{\encoderrnn}$ is the forward LSTM recurrence function. 
We use a similar computation to compute the backward hidden states $\overleftarrow{\hiddenstate_j}$. 
For each token $\inputtoken'_\inputindex$ in $\instruction'$, a vector representation $\hiddenstate_j' = \left[ \overrightarrow{\hiddenstate_j}; \overleftarrow{\hiddenstate_j}\right]$ is computed. 
We then create two sets of vectors, one for all the vectors of the current instruction and one for the previous instructions:

\begin{small}
\begin{eqnarray}
\nonumber \instructionencodings^c &=& \{ \hiddenstate'_\inputindex \}_{\inputindex = \MakeUppercase{\inputindex}}^{\MakeUppercase{\inputindex} + \length{\instruction_\turnindex} }   \\
\nonumber \instructionencodings^p &=& \{ \hiddenstate'_\inputindex \}_{\inputindex = 0}^{\inputindex < \MakeUppercase{\inputindex}}  
\end{eqnarray}
\end{small}

\noindent
where $\MakeUppercase{\inputindex}$ is the index in $\instruction'$ where the current instruction $\instruction_\turnindex$ begins. 
Separating the vectors to two sets will allows computing separate attention on the current instruction and previous ones. 
To compute each input representation during decoding, we use a bi-linear attention function~\cite{Luong:15nmtattention}. 
Given a set of vectors $H$, a query vector $\hiddenstate^q$, and a weight matrix $\mathbf{W}$, the attention function $\func{attend}(H, \hiddenstate^q, \mathbf{W})$ computes a context vector $\contextvector$:

\begin{small}
\begin{eqnarray}
\nonumber \alpha_i &\propto& \exp(\hiddenstate_i^\intercal \mathbf{W} \hiddenstate^q ): i = 0, \dots, \length{H} \\
\nonumber \contextvector &=& \sum_{i=1}^{\length{H}}\alpha_i \hiddenstate_i\;\;.
\end{eqnarray}
\end{small}

We use a decoder to generate actions. 
At each time step $\outputindex$, we compute an input representation using the attention function, update the decoder state, and compute the next action to execute. 
Attention is first computed over the vectors of the current instruction, which is then used to attend over the other inputs. 
We compute the context vectors $\contextvector_\outputindex^c$ and $\contextvector_\outputindex^p$ for the current instruction and previous instructions: 

\begin{small}
\begin{eqnarray}
\nonumber \contextvector^c_\outputindex &=& \func{Attend}(\instructionencodings^c, \hiddenstate^d_{\outputindex-1}, \mathbf{W}^c)	 \\ 
\nonumber \contextvector^p_\outputindex &=& \func{Attend}(\instructionencodings^p, [\hiddenstate^d_{\outputindex-1}, \contextvector^c_\outputindex], \mathbf{W}^p)\;\;,
\end{eqnarray}	
\end{small}

\noindent
where $\hiddenstate^d_{\outputindex-1}$ is the decoder hidden state for step $\outputindex-1$, and $\instructionencodings^c$ and $\instructionencodings^p$ are the sets of vector representations for the current instruction and previous instructions. 
Two attention heads are used over both the initial and current states.  
This allows the model to attend to more than one location in a state at once, for example when transferring items from one beaker to another in \textsc{Alchemy}.
The current state is computed by the transition function $\state_\outputindex = \transition(\state_{\outputindex-1}, \action_{\outputindex-1})$, where $\state_{\outputindex-1}$ and $\action_{\outputindex-1}$ are the state and action at step $\outputindex-1$. 
The context vectors for the initial state $\state_1$ and the current state $\state_\outputindex$ are:

\begin{small}
\begin{eqnarray}
\nonumber	\contextvector^\state_{1,\outputindex} &=& [\func{Attend}( \func{enc}(\state_1), [\hiddenstate^d_{\outputindex-1}, \contextvector^c_\outputindex], \mathbf{W}^{s_b, 1}) ; \\ \nonumber &&~ \func{Attend}(\func{enc}(\state_1), [\hiddenstate^d_{\outputindex-1}, \contextvector^c_\outputindex], \mathbf{W}^{s_b, 2})] \\
\nonumber	\contextvector^{\state}_{\outputindex, \outputindex} &=& [\func{Attend}( \func{enc}(\state_\outputindex), [\hiddenstate^d_{\outputindex-1}, \contextvector^c_\outputindex], \mathbf{W}^{s_c, 1}) ; \\ \nonumber &&~ \func{Attend}(\func{enc}(\state_\outputindex), [\hiddenstate^d_{\outputindex-1}, \contextvector^c_\outputindex], \mathbf{W}^{s_c, 2})]\;\;,
\end{eqnarray}
\end{small}

\noindent
where all $\mathbf{W}^{*,*}$ are learned weight matrices.

We concatenate all computed context vectors with an embedding of the previous action $\action_{\outputindex-1}$ to create the input for the decoder:

\begin{small}
\begin{eqnarray}
\nonumber \hiddenstate_\outputindex &=& \tanh([\contextvector^c_\outputindex ; \contextvector^p_\outputindex; \contextvector^{s}_{1,\outputindex};  \contextvector^{\state}_{\outputindex,\outputindex};  \outputembedding(\action_{\outputindex-1})]\mathbf{W}^d + \mathbf{b}^d) \\
\nonumber \hiddenstate^d_\outputindex &=& \decoderrnn\left(\hiddenstate_\outputindex; \hiddenstate^d_{\outputindex-1}\right)\;\;,
\end{eqnarray}
\end{small}

\noindent
where $\outputembedding$ is a learned action embedding function and $\decoderrnn$ is the LSTM decoder recurrence.

Given the decoder state $\hiddenstate^d_\outputindex$, the next action $\action_\outputindex$ is predicted with a multi-layer perceptron (MLP). 
The actions in our domains decompose to an action type and at most two arguments.\footnote{We use a $\act{NULL}$ argument for unused arguments.} For example, the action $\act{push~1~B}$ in \textsc{Alchemy} has the type $\act{push}$ and two arguments: a beaker number and a color. 
Section~\ref{sec:data} describes the actions of each domain. 
The probability of an action is:

\begin{small}
\begin{eqnarray}
\nonumber \hiddenstate^a_\outputindex &=& \tanh(\hiddenstate^d_\outputindex \mathbf{W}^a) \\
\nonumber 	s_{\outputindex,a_T} &=& \hiddenstate^a_\outputindex \mathbf{b}_{a_T} \\
\nonumber s_{\outputindex,a_1} &=& \hiddenstate^a_\outputindex \mathbf{b}_{a_1} \\
\nonumber s_{\outputindex,a_2} &=& \hiddenstate^a_\outputindex \mathbf{b}_{a_2} \\
\nonumber p(\action_\outputindex = a_T(a_1, a_2) \mid \acontext_\outputindex ; \theta) &\propto & \\ \nonumber && \hspace{-2.5em} \exp(s_{\outputindex, a_T} + s_{\outputindex,a_1} + s_{\outputindex,a_2})\;\;,
\end{eqnarray}
\end{small}

\noindent
where $a_T$, $a_1$, and $a_2$ are an action type, first argument, and second argument. 
If the predicted action is $\act{stop}$, the execution is complete. 
Otherwise, we execute the action $\action_\outputindex$ to generate the next state $\state_{\outputindex+1}$, and update the agent context $\acontext_{\outputindex}$ to $\acontext_{\outputindex+1}$ by appending the pair $(\state_\outputindex,\action_\outputindex)$ to the execution $\execution$ and replacing the current state with $\state_{\outputindex+1}$. 

The model parameters $\theta$ include: the embedding functions $\inembedding$ and $\outputembedding$; the recurrence parameters for $\overrightarrow\encoderrnn$, $\overleftarrow\encoderrnn$, and $\decoderrnn$; $\mathbf{W}^C$, $\mathbf{W}^P$, $\mathbf{W}^{s_b, 1}$, $\mathbf{W}^{s_b, 2}$, $\mathbf{W}^{s_c, 1}$, $\mathbf{W}^{s_c, 2}$, $\mathbf{W}^d$, $\mathbf{W}^a$, and $\mathbf{b}^d$; and the domain dependent parameters, including the parameters of the encoding function $\func{ENC}$ and the action type, first argument, and second argument weights $\mathbf{b}_{a_T}$, $\mathbf{b}_{a_1}$, and $\mathbf{b}_{a_2}$.

\section{Learning}
\label{sec:learning}

We estimate the policy parameters $\theta$ using an exploration-based learning algorithm that  maximizes the immediate expected reward. 
Broadly speaking, during learning, we observe the agent behavior given the current policy, and for each visited state compute the expected immediate reward by observing rewards for all actions. 
We assume access to a set of training examples $\{(\instruction^\dataindex_\turnindex, \state_{\turnindex, 1}^\dataindex, \langle \instruction^\dataindex_1, \dots, \instruction^\dataindex_{\turnindex-1} \rangle, \goalstate^\dataindex_\turnindex) \}_{j=1,\turnindex=1}^{N,\interactionlength^\dataindex}$, where each instruction $\instruction^\dataindex_\turnindex$ is paired with a start state $\state_{\turnindex, 1}^\dataindex$, the previous instructions in the sequence $\langle \instruction^\dataindex_1, \dots, \instruction^\dataindex_{\turnindex-1} \rangle$, and  a goal state $\goalstate^\dataindex_\turnindex$.

\begin{figure}[t!]
\begin{tabular}{@{}p{7cm}}
\vspace{-0.275in}
\begin{algorithm}[H]
\caption{SESTRA: \textbf{S}ingl\textbf{e}-\textbf{st}ep \textbf{R}eward Observ\textbf{a}tion.}\label{alg:learning}
\begin{algorithmic}[1]
\footnotesize
\Require Training data $\{(\instruction^\dataindex_\turnindex, \state_{\turnindex, 1}^\dataindex, \langle \instruction^\dataindex_1, \dots, \instruction^\dataindex_{\turnindex-1} \rangle, $ $\goalstate^\dataindex_\turnindex) \}_{j=1,\turnindex=1}^{N,\interactionlength^\dataindex}$, learning rate $\mu$, entropy regularization coefficient   $\lambda$, episode limit horizon $M$.
\Definitions $\policy_\theta$ is a policy parameterized by $\theta$, $\act{BEG}$ is a special action to use for the first decoder step, and $\act{STOP}$ indicates end of an execution. $\transition(\state, \action)$ is the state transition function, $H$ is an entropy function, $R^\dataindex_\turnindex(\state, \action,\state')$ is the reward function for example $j$ and instruction $\turnindex$, and $\textsc{RMSProp}$ divides each weight by a running average of its squared gradient~\cite{Tieleman:12}.
\Ensure Parameters $\theta$ defining a learned policy $\policy_\theta$.
\For{$t = 1, \dots, T, j = 1, \dots, N$}\label{algline:epoch}
	\For{$i = 1, \dots, n^\dataindex$}
    \State $\execution \leftarrow \langle ~ \rangle, \outputindex \gets 0, \action_0 \gets \mathtt{BEG}$ 
	\State \Comment{Rollout up to $\act{STOP}$ or episode limit.}
    \While{$\action_{\outputindex} \neq \mathtt{STOP} \land \outputindex < M$}\label{algline:rollout}
        \State $\outputindex \leftarrow \outputindex + 1$    	
    	\State $\acontext_\outputindex \gets  (\instruction_\turnindex, \langle \instruction_1, \dots, \instruction_{\turnindex-1}\rangle, \state_\outputindex, \execution[:\outputindex])$
		\State \Comment{Sample an action from policy.}
        \State $\action_\outputindex \sim \policy_\theta (\acontext_\outputindex, \cdot)$
        \State $\state_{\outputindex+1} \leftarrow \transition(\state_\outputindex, \action_\outputindex)$
        \State $\execution \leftarrow [ \execution ;  \langle (\state_\outputindex, \action_\outputindex)\rangle ]$\label{algline:rolloutend}
    \EndWhile
    \State $\Delta \gets \bar{0}$
    \For{$k' = 1,\dots,k$}
        \State \Comment{Compute the entropy of $\pi_\theta(\acontext_{k'}, \cdot)$.}
        \State $\Delta \gets \Delta + \lambda \nabla_\theta  H(\policy_\theta(\acontext_{k'}, \cdot))$ \label{algline:entropy}
        \For{$\action \in \actions$}
        	\State $\state' \gets \transition(\state_{k'}, \action)$
			\State \Comment{Compute gradient for action $\action$.}
            \State $\Delta \leftarrow \Delta + R^\dataindex_\turnindex(\state_{k'}, \action, \state')\nabla_\theta \policy_\theta(\acontext_{k'}, \action)$ \label{algline:action}
        \EndFor
    \EndFor
    \State $\theta \leftarrow \theta + \mu \textsc{RMSProp}\left(\dfrac{\Delta}{\outputindex}\right)$ \label{algline:update}
	\EndFor
\EndFor
\State \Return $\theta$
\end{algorithmic}
\label{alg:learn} 
\end{algorithm}
\end{tabular}
\vspace{-15pt}
\end{figure}

\paragraph{Reward}
The reward $R^\dataindex_\turnindex : \states \times \states \times \actions \rightarrow \mathbb{R}$ is defined for each example $j$ and instruction $\turnindex$: 

\begin{small}
\begin{equation*}
R^\dataindex_\turnindex(\state, \action, \state') = P^\dataindex_\turnindex(\state, \action, \state') + \phi^\dataindex_\turnindex(\state') - \phi^\dataindex_\turnindex(\state)\;\;,
\end{equation*}
\end{small}

\noindent
where $\state$ is a source state, $\action$ is an action, and $\state'$ is a target state.\footnote{While the reward function is defined for any state-action-state tuple, in practice, it is used during learning with tuples that follow the system dynamics, $\state' = T(\state, \action)$.} $P^\dataindex_\turnindex(\state, \action, \state')$ is a problem reward and \mbox{$\phi^\dataindex_\turnindex(\state') - \phi^\dataindex_\turnindex(\state)$} is a shaping term. 
The problem reward $P^\dataindex_\turnindex(\state, \action, \state')$ is positive for stopping at the goal $\goalstate^\dataindex_\turnindex$ and negative for stopping in an incorrect state or taking an invalid action: 

\begin{small}
\begin{equation}
\nonumber P^\dataindex_\turnindex(\state, \action, \state') = 
\begin{cases}
1.0 & \action = \mathtt{STOP} \land \state' = \goalstate^\dataindex_\turnindex \\
-1.0 & \action = \mathtt{STOP} \land \state' \neq \goalstate^\dataindex_\turnindex \\
-1.0 - \delta & \state = \state' \\
-\delta & \text{otherwise}
\end{cases}\;\;,
\end{equation}
\end{small}

\noindent 
where $\delta$ is a verbosity penalty. 
The case $\state = \state'$ indicates that $\action$ was invalid in state $\state$, as in this domain, all valid actions except $\mathtt{STOP}$ modify the state.
We use a potential-based shaping term \mbox{$\phi^\dataindex_\turnindex(\state') - \phi^\dataindex_\turnindex(\state)$}~\cite{Ng:99rewardshaping}, where $\phi^\dataindex_\turnindex(\state) = - || \state - \goalstate^\dataindex_\turnindex||$ computes the edit distance between the state $\state$ and the goal, measured over the objects in each state. 
The shaping term densifies the reward, providing a meaningful signal for learning in nonterminal states.

\paragraph{Objective}
We maximize the immediate expected reward over all actions and use entropy regularization. 
The gradient is approximated by sampling an execution $\execution =  \langle (\state_1, \action_1), \dots, (\state_\outputindex, \action_\outputindex) \rangle$ using our current policy:

\begin{small}
\begin{eqnarray}
\nonumber \nabla_\theta \mathcal{J} &=& \dfrac{1}{\outputindex}\sum_{\outputindex'=1}^{\outputindex}\biggl( \sum_{\action \in \actions}R\left(\state_\outputindex, \action, T(\state_\outputindex, \action)\right)\nabla_\theta\policy(\acontext_\outputindex, \action ) \\ 
\nonumber && + \lambda \nabla_\theta H(\pi(\acontext_\outputindex, \cdot))\biggr)\;\;,
\end{eqnarray}
\end{small}

\noindent
where $H(\pi(\acontext_\outputindex, \cdot)$ is the entropy term.

\paragraph{Algorithm}

Algorithm~\ref{alg:learning} shows the Single-step Reward Observation (\textsc{SESTRA}) learning algorithm. 
We iterate over the training data $T$ times (line~\ref{algline:epoch}). 
For each example $j$ and turn $\turnindex$, we first perform a rollout by sampling an execution $\execution$ from $\pi_\theta$ with at most $M$ actions (lines~\ref{algline:rollout}-\ref{algline:rolloutend}). 
If the rollout reaches the horizon without predicting $\act{stop}$, we set the problem reward $P^{\dataindex}_{\turnindex}$ to $-1.0$ for the last step.
Given the sampled states visited, we compute the entropy (line~\ref{algline:entropy}) and observe the immediate reward for all actions (line~\ref{algline:action}) for each step. 
Entropy and rewards are used to accumulate the gradient, which is applied to the parameters using \textsc{RMSProp}~\cite{Dauphin:15} (line~\ref{algline:update}).

\paragraph{Discussion}

Observing the rewards for all actions for each visited state addresses an on-policy learning exploration problem. Actions that consistently receive negative reward early during learning will be visited with very low probability later on, and in practice, often not explored at all. 
Because the network is randomly initialized, these early negative rewards are translated into strong general biases that are not grounded well in the observed context. 
Our algorithm exposes the agent to such actions later on when they receive positive rewards even though the agent does not explore them during rollout. 
For example, in \textsc{Alchemy}, $\act{pop}$ actions are sufficient to complete the first steps of good executions. As a result, early during learning, the agent learns a strong bias against $\act{PUSH}$ actions. 
In practice, the agent then will not explore $\act{push}$ actions again. 
In our algorithm, as the agent learns to roll out the correct $\act{pop}$ prefix, it is then exposed to the reward for the first $\act{push}$ even though it likely sampled another $\act{pop}$. 
It then unlearns its bias towards predicting $\act{pop}$.%

\begin{figure}[t]
\fbox{
\centering
\begin{minipage}{0.95\linewidth}
	\begin{center}
	\includegraphics[width=\linewidth,clip,trim=197 319 376 242]{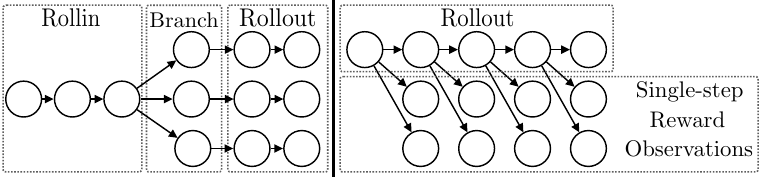}
	\end{center}
\end{minipage}}
\vspace{-5pt}
\caption{Illustration of LOLS~\cite[left; ][]{Chang:15LearningTSbetter} and our learning algorithm (SESTRA, right). LOLS branches a single time, and samples complete rollout for each branch to obtain the trajectory loss.  SESTRA uses a complete on-policy rollout and single-step branching for all actions in each sample state.}
\label{fig:learning}
\vspace{-5pt}	
\end{figure}

Our learning algorithm can be viewed as a cost-sensitive variant of the oracle in \textsc{Dagger}~\cite{Ross:11dagger}, where it provides the rewards for all actions instead of an oracle action. 
It is also related to Locally Optimal Learning to Search~\cite[LOLS;][]{Chang:15LearningTSbetter} with two key distinctions: (a) instead of using different roll-in and roll-out policies, we use the model policy; and (b) we branch at each step, instead of once, but do not rollout from branched actions since we only optimize the immediate reward. Figure~\ref{fig:learning} illustrates the comparison. 
Our summation over immediate rewards for all actions is related the summation  of estimated Q-values for all actions in the Mean Actor-Critic algorithm~\cite{Asadi:17mac}. 
Finally, our approach is related to \citet{Misra:17instructions}, who also maximize the immediate reward, but do not observe rewards for all actions for each state.

\section{SCONE Domains and Data}
\label{sec:data}

\begin{table}[t!]
\centering
\begin{footnotesize}
\begin{tabular}{|l|c|c|c|} \hline
 & \textsc{Alc} & \textsc{Sce} & \textsc{Tan} \\ 
\hline
\# Sequences (train) & $3657$ &  $3352$ & $4189$\\ 
\# Sequences (dev) & $245$ & $198$ & $199$ \\
\# Sequences (test) & $899$ & $1035$ & $800$ \\
\dline
Mean instruction & \multirow{ 2}{*}{\stdev{8.0}{3.2}} & \multirow{ 2}{*}{\stdev{10.5}{5.5}} & \multirow{ 2}{*}{\stdev{5.4}{2.4}} \\ 
~~~length & && \\
Vocabulary size & $695$  & $816$ & $475$\\
\hline
\end{tabular}
\end{footnotesize}
\vspace{-5pt}
\caption{Data statistics for  \textsc{Alchemy} (\textsc{Alc}), \textsc{Scene} (\textsc{Sce}), and \textsc{Tangrams} (\textsc{Tan}).}
\label{tab:stats}
\vspace{-5pt}
\end{table}

\begin{table}[t!]
\begin{footnotesize}
\begin{center}
\begin{tabular}{|l|c|l|c|c|c|c|c} \hline
& Refs/Ex & & $1$ & $2$ & $3$ & $4$  \\ \hline
\multirow{2}{*}{\textsc{Alchemy}} & \multirow{2}{*}{1.4}  & Coref. & $28$ & $7$ & $2$ & $0$  \\ \ddline{3-8}
& & Ellipsis  & $0$ & $0$ & $3$ & $1$  \\ \hline
\multirow{2}{*}{\textsc{Scene}} & \multirow{2}{*}{2.4} & Coref.  & $49$ & $16$ & $5$ & $3$  \\ \ddline{3-8}
& & Ellipsis & $0$ & $0$ & $0$ & $0$  \\ \hline
\multirow{2}{*}{\textsc{Tangrams}} & \multirow{2}{*}{1.7}  & Coref.  & $25$ & $14$ & $2$ & $1$  \\ \ddline{3-8}
& & Ellipsis  & $4$ & $0$ & $0$ & $0$ \\ \hline
\end{tabular}
\end{center}
\end{footnotesize}
\vspace{-5pt}
\caption{Counts of discourse phenomena in SCONE from $30$ randomly selected development interactions for each domain. We count occurrences of coreference between instructions (e.g., \nlstring{he leaves} in \textsc{Scene}) and ellipsis (e.g., \nlstring{then, drain 2 units} in \textsc{Alchemy}), when the last explicit mention of the referent was $1$, $2$, $3$, or $4$ turns in the past. We also report the average number of multi-turn references per interaction  (Refs/Ex).}
\label{tab:analysis}
\vspace{-10pt}
\end{table}

SCONE has three domains: \textsc{Alchemy}, \textsc{Scene}, and \textsc{Tangrams}. 
Each interaction contains five  instructions.
Table~\ref{tab:stats} shows data statistics.
Table~\ref{tab:analysis} shows discourse reference analysis. 
State encodings are detailed in the Supplementary Material. 

\paragraph{\textsc{Alchemy}}
Each environment in \textsc{Alchemy} contains seven numbered beakers, each containing up to four colored chemicals in order. 
Figure~\ref{fig:example} shows an example. 
Instructions describe pouring chemicals between and out of beakers, and mixing beakers.
We treat all beakers as stacks. There are two action types: \act{push} and \act{pop}.
\act{pop} takes a beaker index, and removes the top color. 
\act{push} takes a beaker index and a color, and adds the  color at the top of the beaker. 
To encode a state, we encode each beaker with an RNN, and concatenate the last output with  the beaker index embedding. The set of vectors is the state embedding. 

\paragraph{\textsc{Scene}}
Each environment in \textsc{Scene} contains ten positions, each containing at most one person  defined by a shirt color and an optional hat color. 
Instructions describe adding or removing people, moving a person to another position, and moving a person's hat to another person.
There are four action types: \act{add\_person}, \act{add\_hat}, \act{remove\_person}, and \act{remove\_hat}.
\act{add\_person} and \act{add\_hat} take a position to place the person or hat and the color of the person's shirt or hat.
\act{remove\_person} and \act{remove\_hat} take the position to remove a person or hat from. 
To encode a state, we use a bidirectional RNN over the ordered positions. The input for each position is a concatenation of the color embeddings for the person and hat. 
The set of RNN hidden states is the state embedding. 

\paragraph{\textsc{Tangrams}}
Each environment in \textsc{Tangrams} is a list containing at most five unique objects.
Instructions describe removing or inserting an object into a position in the list, or swapping the positions of two items. 
There are two action types: \act{insert} and \act{remove}. 
\act{insert} takes the position to insert an object, and the  object identifier. 
\act{remove} takes an object position.
We embed each object by concatenating embeddings for its type and position. The resulting set is the state embedding.

\section{Experimental Setup}
\label{sec:exp}

\paragraph{Evaluation}
Following \citet{Long:16context}, we evaluate task completion accuracy using exact match between the final state and the annotated goal state. 
We report accuracy for complete interactions (5utts), the first three utterances of each interaction (3utts), and single instructions (Inst). For single instructions, execution starts from the annotated start state of the instruction. 

\paragraph{Systems}

We report performance of ablations and two baseline systems: \textsc{PolicyGradient}: policy gradient with cumulative episodic reward without a baseline, and \textsc{ContextualBandit}: the contextual bandit approach of \citet{Misra:17instructions}. Both systems use the reward with the shaping term and our model. 
We also report supervised learning results (\textsc{Supervised}) by heuristically generating correct executions and computing maximum-likelihood estimate using  context-action demonstration pairs. Only the supervised approach uses the heuristically generated labels. 
Although the results are not comparable, we also report the performance of previous approaches to SCONE. 
All three approaches generate logical representations based on lambda calculus.
In contrast to our approach, this requires an ontology of hand built symbols and rules to evaluate the logical forms. 
\citet{Fried:17} uses supervised learning with annotated logical forms.

\paragraph{Training Details}
For test results, we run each experiment five times and report results for the model with best validation interaction accuracy. 
For ablations, we do the same with three experiments. 
We use a batch size of $20$. 
We stop training using a validation set sampled from the training data. 
We hold the validation set constant for each domain for all experiments. 
We use patience over the average reward, and select the best model using interaction-level (5utts) validation accuracy. 
We tune $\lambda$, $\delta$, and $M$ on the development set.
The selected values and other implementation details are described in the Supplementary Material.

\section{Results}
\label{sec:results}

\begin{table*}[t]
\begin{footnotesize}
\begin{center}
\begin{tabular}{|l|c|c|c|c|c|c|c|c|c|} \hline
& \multicolumn{3}{c|}{\textsc{Alchemy}} & \multicolumn{3}{c|}{\textsc{Scene}} & \multicolumn{3}{c|}{\textsc{Tangrams}} \\ \cline{2-10}
System & Inst & 3utts & 5utts & Inst & 3utts & 5utts & Inst & 3utts & 5utts \\ \hline\hline
\citet{Long:16context} & -- & $56.8$ & $52.3$ & --& $23.2$ & $14.7$ & --  & $64.9$ & $27.6$  \\ \dline
\citet{Guu:17rl-mml} & -- & $66.9$ & $52.9$ & -- & $64.8$ & $46.2$  & -- & $65.8$  & $37.1$\\ \hline\hline
\citet{Fried:17} & -- & -- & $72.0$ & -- & -- & $72.7$ & -- & -- & $69.6$ \\ \hline\hline
\textsc{Supervised} & $89.4$ & $73.3$ & $62.3$ & $88.8$ & $78.9$ & $66.4$ & $86.6$ & $81.4$ & $60.1$ \\ \hline\hline
\textsc{PolicyGradient} & $0.0$ & $0.0$ & $0.0$ & $0.0$ & $1.3$ & $0.2$ & $84.1$ & $77.4$ & $54.9$\\ \dline
\textsc{ContextualBandit} & $73.8$ & $36.0$ & $25.7$ & $15.1$ & $2.9$ & $4.4$ & $84.8$ & $76.9$ &  $57.9$ \\ \dline
Our approach & $89.1$ & $74.2$ & $62.7$ & $87.1$ & $73.9$ & $62.0$ & $86.6$ & $80.8$ & $62.4$ \\ \hline
\end{tabular}
\end{center}
\end{footnotesize}
\vspace{-5pt}
\caption{Test accuracies for single instructions (Inst), first-three instructions (3utts), and full interactions (5utts).}
\vspace{-3pt}
\label{tab:test}
\end{table*}

\newcommand{\nstdev}[2]{${#1}$ \vspace{-3pt}\newline \hspace*{8pt}\begin{tiny}$\pm {#2}$ \end{tiny}}
\newcommand{\colwidth}{0.8cm}
\begin{table*}[t]
\begin{footnotesize}
\begin{center}
\begin{tabular}{|p{3.2cm}|>{\centering\arraybackslash}p{\colwidth}|>{\centering\arraybackslash}p{\colwidth}|>{\centering\arraybackslash}p{\colwidth}|>{\centering\arraybackslash}p{\colwidth}|>{\centering\arraybackslash}p{\colwidth}|>{\centering\arraybackslash}p{\colwidth}|>{\centering\arraybackslash}p{\colwidth}|>{\centering\arraybackslash}p{\colwidth}|>{\centering\arraybackslash}p{\colwidth}|} \hline
& \multicolumn{3}{c|}{\textsc{Alchemy}} & \multicolumn{3}{c|}{\textsc{Scene}} & \multicolumn{3}{c|}{\textsc{Tangrams}} \\ \cline{2-10}
System & Inst & 3utts & 5utts & Inst & 3utts & 5utts & Inst & 3utts & 5utts \\ \hline\hline
\textsc{Supervised} & $92.0$ & $83.3$ & $71.4$ & $85.3$ & $72.7$ & $60.6$ & $86.1$ & $81.9$ & $58.3$ \\ \hline \hline
\textsc{PolicyGradient} & $0.0$ & $0.0$ & $0.0$ & $0.9$ & $1.0$ & $0.5$ & $85.2$ & $74.9$ & $52.3$\\ \dline
\textsc{ContextualBandit} & $58.8$ & $6.9$ & $5.7$ & $12.0$ & $0.5$ & $1.5$ & $85.6$ & $78.4$ & $52.6$ \\ \hline\hline
Our approach & $\mathbf{92.1}$ & $\mathbf{82.9}$ & $\mathbf{71.8}$ & $\mathbf{83.9}$ & $\mathbf{68.7}$ & $56.1$ & $88.5$ & $82.4$ & $60.3$ \\ \dline
-- previous instructions & $90.1$ & $77.1$ & $66.1$ & $79.3$ & $60.6$ & $45.5$ & $76.4$ & $55.8$ & $27.6$\\ \dline
-- current and initial state & $25.7$ &$4.5$ &$3.3$ & $17.5$ & $0.0$ & $0.0$ & $45.4$ & $15.1$ & $3.5$ \\ \dline
-- current state  & $89.8$ & $78.0$ & $62.9$ & $83.0$ & $\mathbf{68.7}$ & $54.0$ & $87.6$ & $78.4$ & $60.8$ \\ \dline
-- initial state & $81.1$ & $68.6$ & $42.9$ & $82.7$ & $67.7$ & $\mathbf{57.1}$ & $\mathbf{88.6}$ & $\mathbf{82.9}$ & $\mathbf{63.3}$ \\ \hline\hline
Our approach ($\mu \pm \sigma$) & \nstdev{91.5}{1.4} & \nstdev{80.4}{2.6} & \nstdev{69.5}{5.0} & \nstdev{62.9}{17.7} & \nstdev{37.8}{23.5} & \nstdev{29.0}{21.1}  &\nstdev{88.2}{0.6} & \nstdev{80.8}{2.8} & \nstdev{59.2}{2.3}\\ \hline 
\end{tabular}
\end{center}
\end{footnotesize}
\vspace{-5pt}
\caption{Development results, including model ablations. We also report mean $\mu$ and standard deviation $\sigma$ for all metrics for our approach across five experiments. We bold the best performing variations of our model.}
\label{tab:ablations}
\vspace{-6pt}
\end{table*}

\definecolor{myblue}{RGB}{16,132,255}
\definecolor{myyellow}{RGB}{217,209,14}
\definecolor{mylightgrey}{RGB}{198,198,198}
\definecolor{mydarkgrey}{RGB}{95,95,95}

\begin{figure}[t]
\begin{center}
\begin{tikzpicture}
 \begin{axis}[
    width=0.9\columnwidth,
        height=0.5\columnwidth,
        font=\footnotesize,
        xmin=0,xmax=200,
        ymin=0, ymax=1,
        xtick={0,25,50,75,100,125,150,175,200},
        ytick={0,0.25,0.5,0.75,1.0},
        bar width=12pt,
        xlabel style={yshift=1ex,},
        xlabel=\# Epochs,
        ylabel=Accuracy]
             
     \addplot[color=orange,style={thick}] coordinates {
(0,0.0123)
(4,0.1346)
(8,0.1501)
(12,0.2681)
(16,0.2867)
(20,0.3443)
(24,0.7212)
(28,0.7715)
(32,0.7895)
(36,0.7713)
(40,0.7359)
(44,0.7422)
(48,0.7555)
(52,0.7576)
(56,0.7825)
(60,0.7613)
(64,0.7701)
(68,0.7736)
(72,0.7793)
(76,0.7804)
(80,0.7795)
(84,0.7778)
(88,0.8072)
(92,0.7808)
(96,0.7817)
(100,0.8056)
(104,0.8178)
(108,0.8053)
(112,0.8096)
(116,0.8241)
(120,0.8089)
(124,0.7988)
(128,0.7865)
(132,0.7914)
(136,0.7958)
(140,0.7967)
(144,0.7921)
(148,0.7793)
(152,0.8029)
(156,0.7872)
(160,0.7917)
(164,0.7844)
(168,0.7931)
(172,0.7896)
(176,0.7977)
(180,0.8171)
(184,0.7852)
(188,0.8115)
(192,0.7937)
(196,0.8053)
     }; 

     \addplot[color=red,style={thick}] coordinates {     
     (0,0.0104)
(4,0.1355)
(8,0.1501)
(12,0.2842)
(16,0.3375)
(20,0.3843)
(24,0.4081)
(28,0.4178)
(32,0.4251)
(36,0.4346)
(40,0.4373)
(44,0.4359)
(48,0.4419)
(52,0.4403)
(56,0.4437)
(60,0.4454)
(64,0.4491)
(68,0.4496)
(72,0.4502)
(76,0.4517)
(80,0.4510)
(84,0.4535)
(88,0.4557)
(92,0.4551)
(96,0.4545)
(100,0.4553)
(104,0.4582)
(108,0.4591)
(112,0.4625)
(116,0.4641)
(120,0.4637)
(124,0.4623)
(128,0.4657)
(132,0.4672)
(136,0.4646)
(140,0.4682)
(144,0.4676)
(148,0.4686)
(152,0.4700)
(156,0.4691)
(160,0.4698)
(164,0.4705)
(168,0.4711)
(172,0.4727)
(176,0.4709)
(180,0.4707)
(184,0.4727)
(188,0.4742)
(192,0.4722)
(196,0.4718)
};

     \addplot[color=black,style={thick}] coordinates {     
(0,0.0130)
(4,0.1355)
(8,0.2266)
(12,0.2920)
(16,0.3084)
(20,0.3307)
(24,0.3533)
(28,0.3905)
(32,0.7201)
(36,0.7571)
(40,0.7734)
(44,0.7775)
(48,0.7886)
(52,0.7947)
(56,0.7900)
(60,0.8727)
(64,0.8851)
(68,0.8948)
(72,0.8973)
(76,0.9028)
(80,0.9071)
(84,0.9070)
(88,0.9104)
(92,0.9119)
(96,0.9144)
(100,0.9137)
(104,0.8799)
(108,0.9190)
(112,0.9174)
(116,0.9201)
(120,0.9252)
(124,0.9251)
(128,0.9253)
(132,0.9257)
(136,0.9258)
(140,0.9290)
(144,0.9264)
(148,0.9285)
(152,0.9258)
(156,0.9314)
(160,0.9309)
(164,0.9318)
(168,0.9353)
(172,0.9352)
(176,0.9379)
(180,0.9363)
(184,0.9368)
(188,0.9395)
(192,0.9400)
(196,0.9386)
}; 

     \addplot[color=mylightgrey,style={thick}] coordinates {     
(0,0.0108)
(4,0.1343)
(8,0.1477)
(12,0.2999)
(16,0.3301)
(20,0.3541)
(24,0.5080)
(28,0.5208)
(32,0.5287)
(36,0.5255)
(40,0.5339)
(44,0.5360)
(48,0.5390)
(52,0.5371)
(56,0.5412)
(60,0.5418)
(64,0.5412)
(68,0.5436)
(72,0.5458)
(76,0.5436)
(80,0.5455)
(84,0.5491)
(88,0.5475)
(92,0.5481)
(96,0.5489)
(100,0.5494)
(104,0.5518)
(108,0.5521)
(112,0.5519)
(116,0.5528)
(120,0.5554)
(124,0.5525)
(128,0.5538)
(132,0.5556)
(136,0.5547)
(140,0.5565)
(144,0.5550)
(148,0.5554)
(152,0.5580)
(156,0.5568)
(160,0.5588)
(164,0.5579)
(168,0.5590)
(172,0.5583)
(176,0.5601)};

     \addplot[color=myblue,style={thick}] coordinates {     
(0,0.0128)
(4,0.1389)
(8,0.1747)
(12,0.2695)
(16,0.3147)
(20,0.3559)
(24,0.3811)
(28,0.4106)
(32,0.4299)
(36,0.4357)
(40,0.4394)
(44,0.4385)
(48,0.4383)
(52,0.4437)
(56,0.4476)
(60,0.4471)
(64,0.4500)
(68,0.4475)
(72,0.4550)
(76,0.4505)
(80,0.4557)
(84,0.4548)
(88,0.4560)
(92,0.4552)
(96,0.4578)
(100,0.4578)
(104,0.4580)
(108,0.4563)
(112,0.4607)
(116,0.4611)
(120,0.4621)
(124,0.4617)
(128,0.4631)
(132,0.4621)
(136,0.4640)
(140,0.4640)
(144,0.4614)
(148,0.4633)
(152,0.4643)
(156,0.4622)
(160,0.4655)
(164,0.4657)
(168,0.4670)
(172,0.4671)
(176,0.4666)
(180,0.4676)
(184,0.4688)
(188,0.4673)
(192,0.4704)
(196,0.4707)
};
     \end{axis}
\end{tikzpicture}
\end{center}
\vspace{-10pt}
\caption{Instruction-level training accuracy per epoch when training five models on \textsc{Scene}, demonstrating the effect of randomization in the learning method. Three of five experiments fail to learn effective models. The red and blue learning trajectories are overlapping.}
\label{fig:scenelearning}
\vspace{-15pt}
\end{figure}
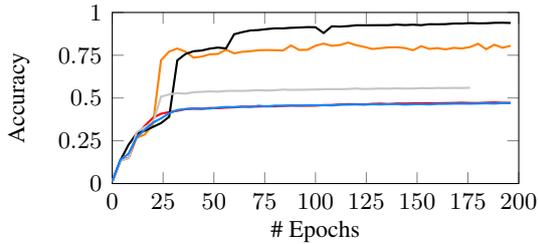  

Table~\ref{tab:test} shows test results. 
Our approach significantly outperforms \textsc{PolicyGradient} and \textsc{ContextualBandit}, both of which suffer due to biases learned early during learning, hindering later exploration. 
This problem does not appear in \textsc{Tangrams}, where no action type is dominant at the beginning of executions, and all methods perform well. 
\textsc{PolicyGradient} completely fails to learn \textsc{Alchemy} and \textsc{Scene} due to observing only negative total rewards early during learning. Using a baseline, for example with an actor-critic method, will potentially close the gap to \textsc{ContextualBandit}. However, it is unlikely to address the on-policy exploration problem.

Table~\ref{tab:ablations} shows development results, including model ablation studies. 
Removing previous instructions (-- previous instructions) or both states (-- current and initial state) reduces performance across all domains. 
Removing only the initial state (-- initial state) or the current state (-- current state) shows mixed results across the domains. 
Providing access to both initial and current states increases performance for \textsc{Alchemy}, but reduces performance on the other domains. We hypothesize that this is due to the increase in the number of parameters outweighing what is relatively marginal information for these domains. 
In our development and test results we use a single architecture across the three domains, the full approach, which has the highest interactive-level accuracy when averaged across the three domains ($62.7$ 5utts).
We also report mean and standard deviation for our approach over five trials.
We observe exceptionally high variance in performance on \textsc{Scene}, where some experiments fail to learn and training performance remains exceptionally low (Figure~\ref{fig:scenelearning}). This highlights the sensitivity of the model to the random effects of initialization, dropout, and ordering of training examples.

\begin{table}
\begin{footnotesize}
\begin{center}
\begin{tabular}{|p{4cm}|c|c|c|} \hline
Class & \textsc{Alc} & \textsc{Sce} & \textsc{Tan} \\ \hline
State reference & $23$ & $13$ & $7$ \\ \dline %
Multi-turn reference & $12$ & $5$ & $13$ \\ \dline %
Impossible multi-turn reference & $2$ & $5$ & $13$ \\ \dline %
Ambiguous or incorrect label & $2$ & $19$ & $12$ \\ \hline %
\end{tabular}
\end{center}
\end{footnotesize}
\vspace{-5pt}
\caption{Common error counts in the three domains.}
\label{tab:error_analysis}
\vspace{-5pt}
\end{table}

We analyze the instruction-level errors made by our best models when the agent is provided the correct initial state for the instruction. 
We study fifty examples in each domain to identify the type of failures. 
Table~\ref{tab:error_analysis} shows the counts of major error categories.
We consider multiple reference resolution errors. 
State reference errors indicate a failure to resolve a reference to the world state.
For example, in \textsc{Alchemy}, the phrase \nlstring{leftmost red beaker} specifies a beaker in the environment. If the model picked the correct action, but the wrong beaker, we count it as a state reference. 
We distinguish between multi-turn reference errors that should be feasible, and these that that are impossible to solve without access to states before executing previous utterances, which are not provided to our model. 
For example, in \textsc{Tangrams}, the instruction \nlstring{put it back in the same place} refers to a previously-removed item.
Because the agent only has access to the world state after following this instruction, it does not observe what kind of item was previously removed, and  cannot identify the item to add. 
We also find a significant number of errors due to ambiguous or incorrect instructions.
For example, the \textsc{Scene} instruction \nlstring{person in green appears on the right end} is ambiguous.
In the annotated goal, it is interpreted as referring to a person already in the environment, who moves to the 10th position.
However, it can also be interpreted as a new person in green appearing in the 10th position. 

We also study  performance with respect to multi-turn coreference by observing whether the model was able to identify the correct referent for each occurrence included in the analysis in Table~\ref{tab:analysis}. 
The models were able to correctly resolve $92.3\%$, $88.7\%$, and $76.0\%$ of references in \textsc{Alchemy}, \textsc{Scene}, and \textsc{Tangrams} respectively.

Finally, we include attention visualization for examples from the three domains in the Supplementary Material.

\section{Discussion}
\label{sec:discuss}

We propose a model to reason about context-dependent instructional language that display strong dependencies both on the history of the interaction and the state of the world. 
Future modeling work may include using intermediate world states from previous turns in the interaction, which is required for some of the most complex references in the data.
We propose to train our model using SESTRA, a learning algorithm that takes advantage of single-step reward observations to overcome learned biases in on-policy learning. 
Our learning approach requires additional reward observations in comparison to conventional reinforcement learning. 
However, it is particularly suitable to recovering from biases acquired early during learning, for example due to biased action spaces, which is likely to lead to incorrect blame assignment in neural network policies. 
When the domain and model are less susceptible to such biases, the benefit of the additional reward observations is less pronounced.
One possible direction for future work is to use an estimator to predict rewards for all actions, rather than observing them. 

\section*{Acknowledgements}
This research was supported by the NSF (CRII-1656998), Schmidt Sciences, and cloud computing credits from Amazon.
We thank John Langford and Dipendra Misra for helpful and insightful discussions with regards to our learning algorithm. We also thank the anonymous reviewers for their helpful comments.

\balance
\bibliography{main}
\bibliographystyle{acl_natbib}

\clearpage

\nobalance

\appendix

\section{Domain-Specific Implementation Details}
\label{sec:sup:domains}

For each domain \textsc{Alchemy}, \textsc{Scene}, and \textsc{Tangrams}, we describe the world state representation, state distance function, transition function, and the state encoder.
For all states $\state$, $\state = \transition(\state, \mathtt{STOP})$.

\paragraph{\textsc{Alchemy}}
The world state in \textsc{Alchemy} is a sequence of beakers $\langle \beaker_1, \beaker_2, ..., \beaker_\alchemylength \rangle$ of fixed length $\alchemylength = 7$.
Each beaker $\beaker_\beakerindex = \langle \beakercolor_{\beakerindex, 1}, \beakercolor_{\beakerindex, 2}, ... \beakercolor_{\beakerindex, \length{\beaker_\beakerindex}} \rangle$ is a variable length sequence containing chemical units $\beakercolor$, each one of six possible colors.
The distance between two world states is the sum over distances for each corresponding beaker pair. The distance between two beakers is the edit distance of the list of chemical units in each.
The action space of \textsc{Alchemy} includes two action types, $\act{pop}$ and $\act{push}$.
The $\act{pop}$ action takes one argument: $\act{n} \in \{1, \dots, N\}$ denoting the beaker to pop a chemical unit from.
The $\act{push}$ action takes two arguments: $\act{n}$ and $\act{c}$, one of six colors.
The transition function $\transition$ is defined by two cases: (a) $\transition(\state, \action = \act{push~ n~ c})$ will return a state where $\act{c}$ is added to the beaker with index $\act{n}$; and (b) $\transition(\state, \action = \act{pop~ n~})$ will remove the top element from the beaker with index $\act{n}$, or if the beaker with index $\act{n}$ is empty, the input state $\state$ is returned. 
The state encoding function $\func{Enc}$ is parameterized by (a) $\unitembedding$, an embedding function for each color; (b) $\posembedding$, a positional embedding function for each beaker position; and (c) $\beakerrnn$, a forward RNN used to encode each beaker. 
We encoder each beaker $\beaker_\beakerindex$ with an RNN:

\begin{small}
\begin{equation*}
\hiddenstate^b_{\beakerindex, j} = \beakerrnn\left(\unitembedding(\beakercolor_{\beakerindex, j}); \hiddenstate^b_{\beakerindex, j-1}\right)\;\;.
\end{equation*}
\end{small}

\noindent $\func{ENC}$  returns a set of $N$ vectors $\{ \hiddenstate_\beakerindex \}_{\beakerindex=1}^N$ , where each $\hiddenstate_\beakerindex = [\hiddenstate^b_{\beakerindex, \length{\beaker_\beakerindex}}; \posembedding(\beakerindex)]$ represents a beaker.

\paragraph{\textsc{Scene}}
The world state in \textsc{Scene} is a sequence of positions $\scenestate = \langle \spot_1, \spot_2, ..., \spot_\scenelength \rangle$ of fixed length $\scenelength=10$. 
Each position is a tuple $\spot_\spotindex = \langle \shirtcolor_\spotindex, \hatcolor_\spotindex \rangle$, where $\shirtcolor_\spotindex$ is a shirt color $\hatcolor_\spotindex$ is a hat color. There are six colors, and a special $\mathtt{NULL}$ marker indicating no shirt or hat is present. 
The distance between two world states is the sum over positions of the number of steps required to modify two corresponding positions to be the same given the domain actions space.
The action space of \textsc{Scene} includes four action types: $\act{appear\_person}$, $\act{appear\_hat}$, $\act{remove\_person}$, and $\act{remove\_hat}$.
$\act{appear\_person}$ and $\act{appear\_hat}$ take two arguments: a position index $\act{N}$  and a color $\act{C}$.
$\act{remove\_person}$ and $\act{remove\_hat}$ take one argument: a position index $\act{N}$. 
The transition function $\transition$ is defined by four cases: (a) $\transition(\state, \action = \act{appear\_person~ N~ C})$ returns a state where position $\act{N}$ contains shirt color $\act{C}$ if the shirt color in position $\act{N}$ is $\mathtt{NULL}$, otherwise the action is invalid and the input state $\state$ is returned; (b) $\transition(\state, \action = \act{appear\_hat~ N~ C})$ is defined analogously to $\act{appear\_person}$; (c) $T(\state, \action = \act{remove\_person~ N})$ returns a state where the shirt color at position $\act{N}$ is set to $\mathtt{NULL}$ if there is a color at position $\act{N}$, otherwise the action is invalid and the input state $\state$ is returned; and (d) $T(\state, \action = \act{remove\_hat~ N})$ is defined analogously to $\act{remove\_person}$. 
The state encoding function $\func{Enc}$ is parameterized by (a) $\unitembedding$, an embedding function for shirt and hat colors; (b) $\posembedding$, a positional embedding for each position in the scene; and (c) $\scenernn$, a bidirectional RNN over all positions in order.
Each position is embedded using a function $\phi'(\spot_\spotindex) = [\unitembedding(s_\spotindex); \unitembedding(h_\spotindex); \posembedding(\spotindex)]$.
We compute a sequence of forward hidden states: 

\begin{small}
\begin{equation*}
\overrightarrow{\hiddenstate}^s_{\spotindex} = \overrightarrow{\scenernn}\left(\phi'(\spot_\spotindex); \overrightarrow{\hiddenstate}^s_{\spotindex -1}\right)\;\;.
\end{equation*}
\end{small}

\noindent The backward RNN is equivalent.  $\func{ENC}$ returns the set $\{ \hiddenstate_\spotindex\}_{\spotindex = 1}^N$, where $\hiddenstate_\spotindex = [ \overrightarrow{\hiddenstate}^s_{\spotindex} ; \overleftarrow{\hiddenstate}^s_{\spotindex}; \phi'(\spot_\spotindex)]$ represents a position.

\paragraph{\textsc{Tangrams}}
The world state in \textsc{Tangrams} is a list of positions $\tangramstate = \langle \spot_1, \spot_2, ..., \spot_\tangramlength \rangle$ of a variable length $\tangramlength$. 
Each position contains one of five unique shapes.
The distance function between states is the edit distance between the lists, with a cost of two for substitutions.
The action space of \textsc{Tangrams} includes two action types, $\act{INSERT}$ and $\act{REMOVE}$. 
The $\act{INSERT}$ action takes two arguments: a position $\act{N} \in \{ 1, \cdots, M\}$, where $M$ is the maximum length of a state in the \textsc{Tangrams} dataset, and a shape type $\act{T}$, which is one the five possible shapes.
The $\act{REMOVE}$ action takes a single argument: a position $\act{N}$. 
The transition function $\transition$ is defined by two cases: (a) $\transition(\state, \action = \act{INSERT~ N~ T})$ returns a state where the shape $\act{T}$ is in position $\act{N}$ and all objects to its right shifted by one position if $\act{T}$ is not already in the state, otherwise the action is invalid and $\state$ is returned; and (b) $\transition(\state, \action = \act{REMOVE~ N})$ returns a state where the object in position $\act{N}$ was removed if $\act{N} \leq \tangramlength$, otherwise the action is invalid and $\state$ is returned. 
The state encoding function $\func{ENC}$ is parameterized by (a) $\hiddenstate_{NULL}$, a vector used when $\tangramlength=0$; (b) $\shapeembedding$, an embedding function for the shapes; and (c) $\posembedding$, a positional embedding of the position $\spotindex$.
$\func{ENC}$ returns a set $\{ \hiddenstate_\spotindex\}_{\spotindex=1}^\tangramlength$, where $\hiddenstate_\spotindex = [\posembedding(\spotindex); \shapeembedding(\spot_\spotindex)]$ is the position encoding, or it returns $\{\hiddenstate_{NULL}\}$ if the state contains no objects.

\section{Data Analysis}
\label{sec:sup:data}

We analyze SCONE to identify the frequency of various discourse phenomena in the three domains, including explicit coreference and ellipsis, which is implicit reference  to previous entities. 
We observe references to previous objects (e.g., beakers in \textsc{Alchemy}), actions, locations (e.g., positions in \textsc{Scene}), and world states.
We analyze thirty development set interactions for each domain for presence of these references. We define the age of each referent as the number of turns since it was last explicitly mentioned.
This illustrates the extent to which this dataset challenges models for context-dependent reasoning.

\begin{table}
\begin{footnotesize}
\begin{center}
\begin{tabular}{|l|l|c|c|c|c|c} \hline
& & $1$ & $2$ & $3$ & $4$  \\ \hline
\multirow{3}{*}{Coref.} & Beaker & $24$ & $7$ & $2$ & $0$ \\ \ddline{2-6}
& Action & $3$ & $0$ & $0$ & $0$\\ \ddline{2-6}
& Action + Arguments & $1$ & $0$ & $0$ & $0$ \\ \hline
\multirow{1}{*}{Ellipsis} & Beaker & $0$ & $0$ & $3$ & $1$ \\ \hline
\end{tabular}
\end{center}
\end{footnotesize}
\caption{Count of phenomena in \textsc{Alchemy}.}
\label{tab:alchemy}
\end{table}

\paragraph{\textsc{Alchemy}} Table~\ref{tab:alchemy} shows phenomena counts in \textsc{Alchemy}. 
Each interaction contains on average $1.4$ references dependent on the interaction history. 
Each non-first utterance contains on average $0.3$ references. 
The most common form of reference is explicit coreference (Coref.) to previously-mentioned beakers, for example \nlstring{mix it}.
Other references are to previous actions, referring to the action only (e.g., \nlstring{same with the last beaker}) or the action as well as the arguments (e.g., \nlstring{same for one more unit}, referring to draining one unit from a previously-used beaker). 
Ellipsis occurred four times in the thirty evaluated interactions, for example \nlstring{then, drain 1 unit}, implicitly referring to a specific beaker to drain from.

\begin{table}
\begin{footnotesize}
\begin{center}
\begin{tabular}{|l|l|c|c|c|c|c} \hline
& & $1$ & $2$ & $3$ & $4$  \\ \hline
\multirow{3}{*}{Coref.} & Person & $42$ & $16$ & $5$ & $3$ \\ \ddline{2-6}
& Hat & $2$ & $0$ & $0$ & $0$ \\ \ddline{2-6}
& Action + Arguments & $3$ & $0$ & $0$ & $0$ \\ \ddline{2-6}
& Position & $2$ & $0$ & $0$ & $0$ \\ \hline
\end{tabular}
\end{center}
\end{footnotesize}
\caption{Count of phenomena in \textsc{Scene}.}
\label{tab:scene}
\end{table}

\paragraph{\textsc{Scene}} Table~\ref{tab:scene} shows phenomena counts in \textsc{Scene}. 
Each interaction contains on average $2.4$ references dependent on the interaction history.
Each non-first utterance contains on average $0.6$ references.
The most common form of reference is explicit coreference (Coref.) to previously-mentioned people, for example \nlstring{he moves to the left end}.
Coreference also occurs on hat colors (e.g., \nlstring{he gives it back}), actions along with their arguments (e.g., \nlstring{they did it again} referring to trading specific hats), and positions (e.g., \nlstring{he moves back}).

\begin{table}
\begin{footnotesize}
\begin{center}
\begin{tabular}{|l|l|c|c|c|c|c} \hline
& & $1$ & $2$ & $3$ & $4$  \\ \hline
\multirow{3}{*}{Coref.} & Object & $13$ & $0$ & $0$ & $0$ \\ \ddline{2-6}
& Object via Arguments & $6$ & $10$ & $2$ & $1$ \\ \ddline{2-6}
& Position & $0$ & $2$ & $0$ & $0$ \\ \ddline{2-6}
& Action & $5$ & $2$ & $0$ & $0$ \\ \ddline{2-6}
& Action + Arguments & $1$ & $0$ & $0$ & $0$ \\ \hline
\multirow{2}{*}{Ellipsis} & Position & $3$ & $0$ & $0$ & $0$ \\ \ddline{2-6}
& Action + Arguments & $1$ & $0$ & $0$ & $0$ \\ \hline
\end{tabular}
\end{center}
\end{footnotesize}
\caption{Count of phenomena in \textsc{Tangrams}.}
\label{tab:tangrams}
\end{table}

\paragraph{\textsc{Tangrams}} Table~\ref{tab:tangrams} shows phenomena counts in \textsc{Tangrams}. 
Each interaction contains on average $1.7$ references dependent on the interaction history.
Each non-first utterance contains on average $0.4$ references.
The most common form of reference is on objects via reference to a previous step, for example \nlstring{put the item you just removed in the second spot}.
This requires recalling actions taken in previous turns, including the actions' arguments and the previous world state.
Coreference (Coref.) also occurs for positions (e.g., \nlstring{...where the last deleted figure was}), actions (e.g., \nlstring{do the same with the second to last figure and one before it}), and actions along with the previously-used arguments (e.g., \nlstring{repeat the first step}).
Ellipsis occurs for positions (e.g., \nlstring{add it again}, implicitly referring to the item's previous location) and actions along with their arguments (e.g., \nlstring{undo the last step}).

\section{Attention Analysis}
\label{sec:sup:attn}

\begin{figure*}[t]
\fbox{
\centering
	\includegraphics[width=\linewidth,clip,trim=10 60 50 35,right]{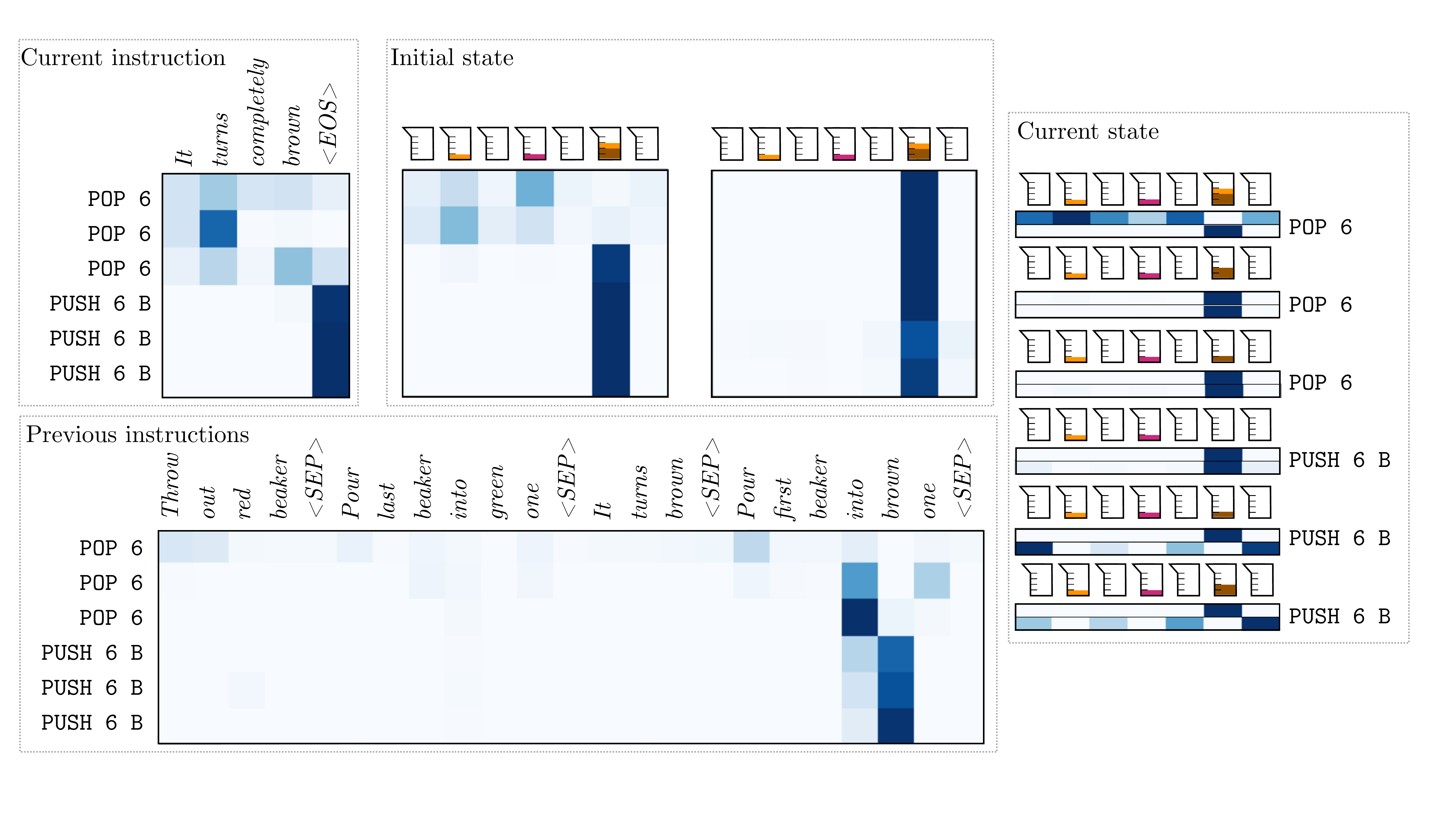}}
\caption{Example of the attention distributions for executing the instruction \nlstring{It turns completely brown} in \textsc{Alchemy}.
This is the fifth instruction in the interaction. 
The correct action sequence mixes the chemicals in the sixth beaker by removing the three units and re-adding three brown units. Our model correctly predicts this sequence. 
We show the different attention distributions when generating this sequence of actions. 
Clockwise starting from the top left: (a) attention over the current instruction; (b) two attention heads over the initial state; (c) two attention heads over the current world state, which changes following each action; and (d) the attention over the previous instructions in the interaction. 
}
\label{fig:attention}	
\end{figure*}

\begin{figure*}[t]
\centering
\fbox{
\centering
	\centering
	\includegraphics[width=\linewidth,clip,trim=0 200 350 70,right]{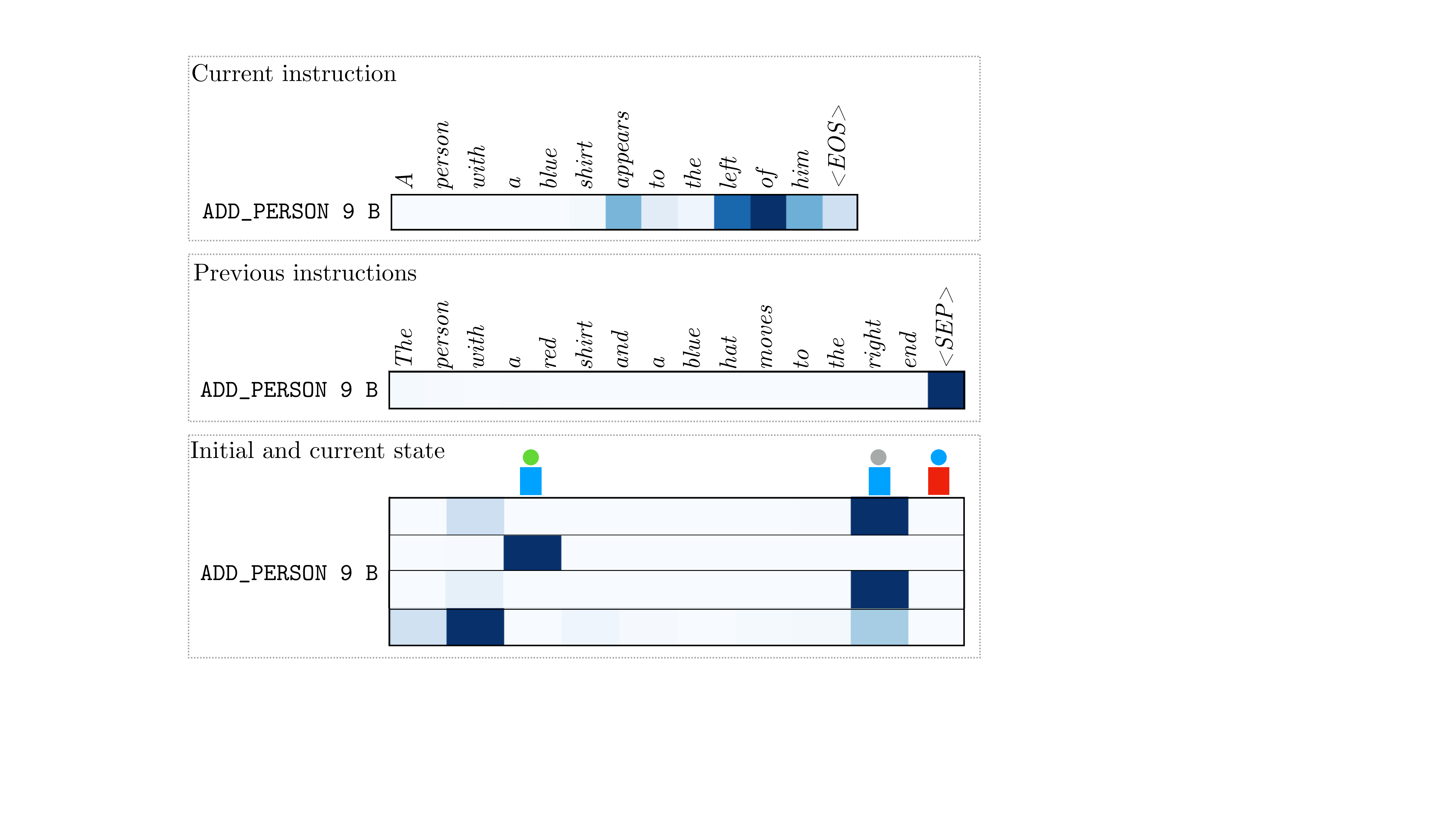} }
\caption{Example of attention for a randomly selected instruction from the development set for \textsc{Scene}.
The instruction \nlstring{A person with a blue shirt appears to the left of him} is the second in the interaction, following the instruction \nlstring{The person with a red shirt and a blue hat moves to the right end}.
The correct action sequence consists of a single action, $\act{ADD\_PERSON\ 9\ B}$, where a person wearing a blue shirt appears in position $9$, to the left of the person in the red shirt. Our model predicts this action correctly.
We show the different attention distributions when generating this sequence of a single action. 
From top to bottom: (a) attention over the current instruction; (b) attention over the previous instruction; and (c) attention over the world state.
As the sequence contains a single action only, the current and initial world states are the same, and their distributions are shown together.
There are two attention heads over both the initial (top two rows) and current (bottom two rows) world states.}
\label{fig:scene}	
\end{figure*}

\begin{figure*}[t]
\fbox{
\centering
	\includegraphics[width=\linewidth,clip,trim=20 175 30 100, right]{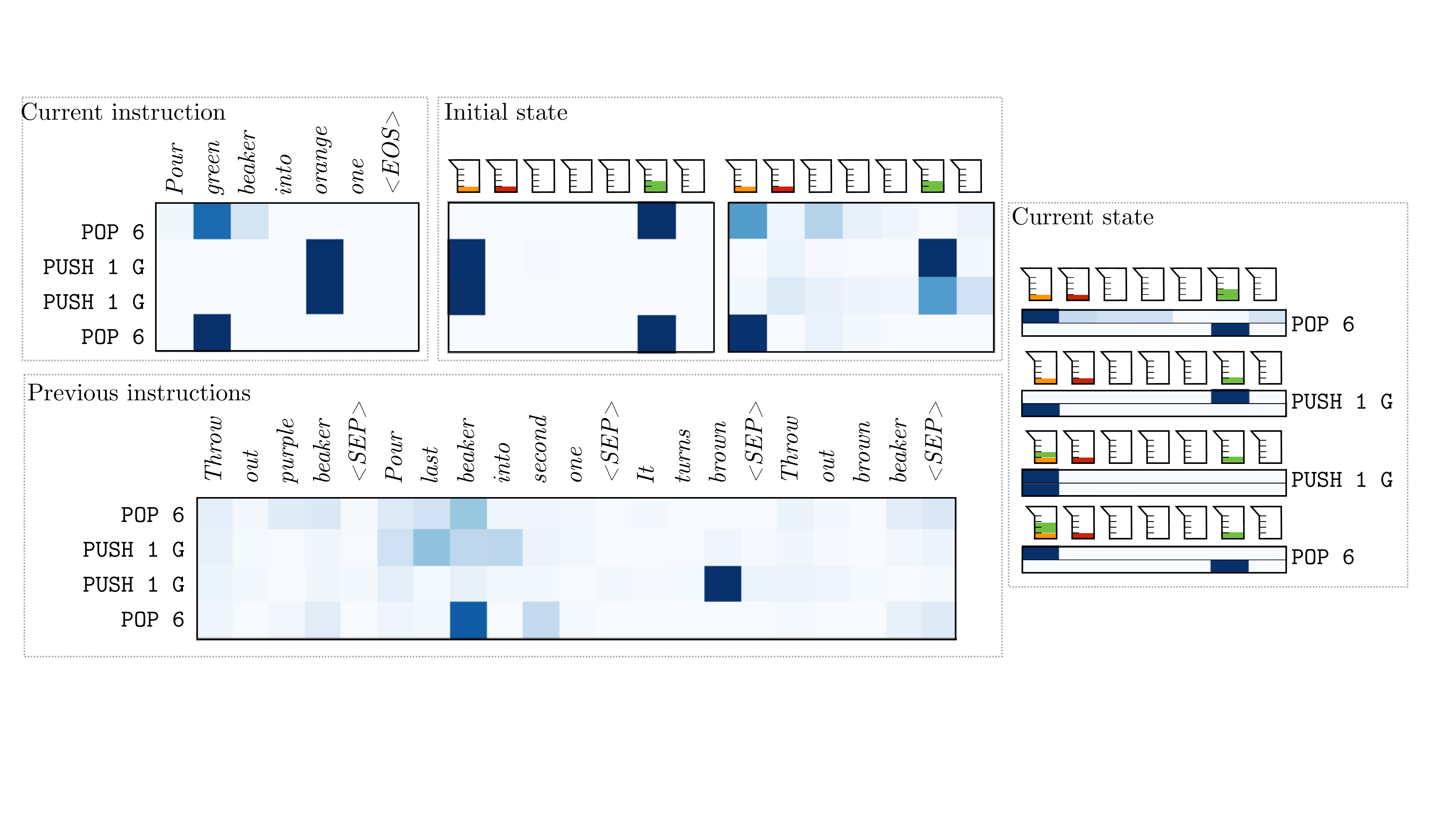}}
\caption{Example of attention for a randomly selected instruction from the development set for \textsc{Alchemy}. The instruction executed is \nlstring{Pour green beaker into orange one}, the fifth instruction in the sequence.  We show the different attention distributions when generating the correct action sequence, which removes green items from the sixth beaker and adds the same number of green items to the beaker containing orange. Clockwise starting from the top left: (a) attention on the current instruction; (b) the two attention heads over the initial state; (c) the two attention heads over the current state as it changes during execution; and (d)  attention over previous instructions.}
\label{fig:alchemy}	
\end{figure*}

\begin{figure*}[t]
\fbox{
\centering
	\includegraphics[width=\linewidth,clip,trim=290 415 170 40, right]{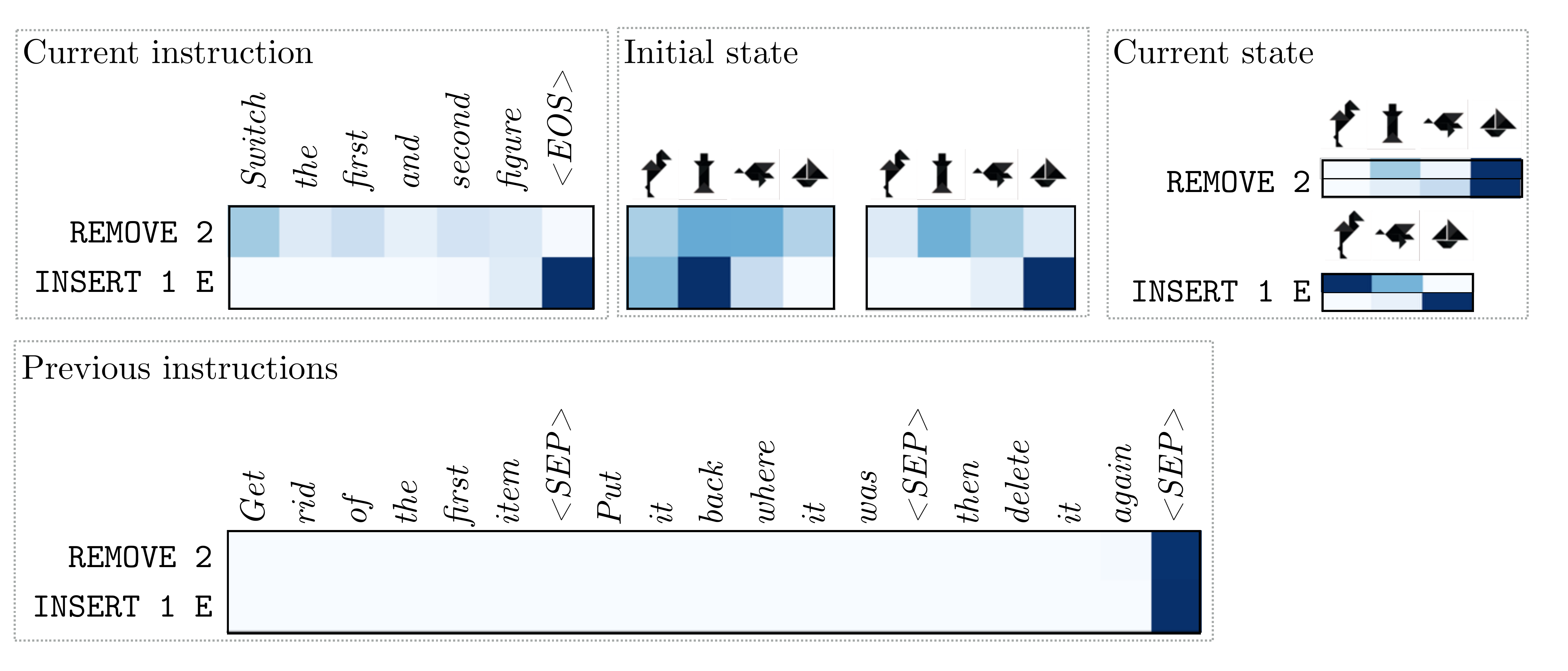} }
\caption{Example of attention for a randomly selected instruction from the development set for \textsc{Tangrams}. The instruction executed is \nlstring{Switch the first and second figure}, the fourth instruction in the sequence.
We show the different attention distributions when generating the correct action sequence, which removes the figure in position two and adds it in position one, thereby swapping the first two items. Clockwise starting from the top left: (a) attention on the current instruction; (b) the two attention heads over the initial state; (c) the two attention heads over the current state as it changes during execution; and (d)  attention over previous instructions.}
\label{fig:tangrams}	
\end{figure*}

Figure~\ref{fig:attention} shows attention distributions for a hand-picked example in \textsc{Alchemy}.
We show the attention probabilities ($\alpha$ in Section~\ref{sec:model}) for the current and previous utterances, initial state, and current state throughout execution.
In this example, the previous-instruction attention puts most of the weight on \nlstring{brown one} during generation, which is the referent of \nlstring{it} in the current instruction.
The initial and current state attentions are placed heavily on the beaker being manipulated.
However, for randomly selected examples, we observe that the attention distribution does not always correspond to intuitions about what should be attended on. 
Figures~~\ref{fig:scene}, \ref{fig:alchemy}, and ~\ref{fig:tangrams} show examples of attention distributions for three random instructions in the development sets of the three domains where the action sequence was predicted correctly.

\section{Architecture and Training Details}
\label{sec:sup:arch}

\paragraph{Model Architecture}
We use an embedding size of $50$ for words and action types and arguments. 
Action embedding is a concatenation of the embeddings of each part, including the action type and the two arguments; an embedded action is a vector of size $150$. 
Embeddings of colors in \textsc{Alchemy} and \textsc{Scene}, and shapes in \textsc{Tangrams}, are of size $10$.
Positional embeddings are of size $10$.
$\mathbf{W}^d$ and $\mathbf{W}^a$ are square matrices. 
All matrices are initialized by sampling from the uniform distribution $U\left(\left[-\sqrt{\frac{6}{M+N}}, \sqrt{\frac{6}{M+N}}\right]\right)$~\cite{Glorot:2010understanding}, where $M$ and $N$ are the matrix dimensionality. 
All RNNs are single-layer LSTMs.
For the main model, both the instruction encoder and action sequence decoder use a hidden size of $100$ in each direction.
The action sequence decoder is initialized by first setting the hidden state and cell memory to zero-vectors, and passing in a zero-vector to update the states, after which attention is computed for the first time.
For \textsc{Alchemy}, the world state encoder has a hidden size of $20$. 
For \textsc{Scene}, the world state encoder has a hidden size of $5$.

\paragraph{Training}
We apply dropout in three places: (a) in each attention computation after multiplying by $\mathbf{W}$; (b) after computing $\mathbf{h}_\outputindex$, the input to each decoder step; and (c) for all attention keys except for the current utterance. 
For \textsc{PolicyGradient}, \textsc{ContextualBandit}, and our approach, we optimize parameters using \textsc{RMSProp}~\cite{Tieleman:12}.
For supervised learning, we use \textsc{Adam}~\cite{Kingma:14adam} for optimization.
We use a learning rate of $0.001$ for all experiments.
Our validation set is a held-out subset containing $7.0\%$ of the training data.
We stop training by observing the instruction-level reward on the validation set.
We use patience for early stopping. We reset patience to $50 \cdot 1.005^x$ the $x$-th time the reward has improved on the validation set, decrease by one each epoch reward does not improve, and stop when patience runs out.
Regardless of patience, we terminate training after $200$ epochs.
We tune $\lambda$, $\delta$, and $M$ on the development set.
In \textsc{Alchemy}, $\lambda = 0.1$, $\delta=0.15$, and $M = 7$.
In \textsc{Scene}, $\lambda = 0.07$, $\delta = 0.2$, and $M = 5$.
In \textsc{Tangrams}, $\lambda=0.1$, $\delta=0.0$, and $M=5$.

\end{document}